\pdfoutput=1

\documentclass[11pt]{article}

\usepackage[preprint]{acl}

\usepackage{times}
\usepackage{latexsym}

\usepackage[T1]{fontenc}

\usepackage[utf8]{inputenc}

\usepackage{microtype}

\usepackage{inconsolata}

\usepackage{graphicx}
\usepackage{subcaption}

\usepackage{amsmath}
\usepackage{amssymb}

\usepackage{multirow}
\usepackage{booktabs}
\usepackage{siunitx}
\usepackage{url}

\usepackage{adjustbox} 

\usepackage{pifont} 

\title{CFiCS: Graph-Based Classification of Common Factors and Microcounseling Skills}

\author{
 \textbf{Fabian Schmidt\textsuperscript{1}},
 \textbf{Karin Hammerfald\textsuperscript{2}},
 \textbf{Henrik Haaland Jahren\textsuperscript{3}},
\textbf{Vladimir Vlassov\textsuperscript{1}}
\\
\\
 \textsuperscript{1}Department of Computer Science, KTH Royal Institute of Technology, Stockholm, Sweden\\
 \textsuperscript{2}Department of Psychology, University of Oslo, Oslo, Norway\\
 \textsuperscript{3}Braive AS, Oslo, Norway
\\
 \small{
   \textbf{Correspondence:} \href{mailto:fschm@kth.se}{fschm@kth.se}
 }
}

\begin{document}
\maketitle
\maketitle
\begin{abstract}
Common factors and microcounseling skills are critical to the effectiveness of psychotherapy. Understanding and measuring these elements provides valuable insights into therapeutic processes and outcomes. However, automatic identification of these change principles from textual data remains challenging due to the nuanced and context-dependent nature of therapeutic dialogue. This paper introduces CFiCS, a hierarchical classification framework integrating graph machine learning with pretrained contextual embeddings. We represent common factors, intervention concepts, and microcounseling skills as a heterogeneous graph, where textual information from ClinicalBERT enriches each node. This structure captures both the hierarchical relationships (e.g., skill-level nodes linking to broad factors) and the semantic properties of therapeutic concepts.
By leveraging graph neural networks, CFiCS learns inductive node embeddings that generalize to unseen text samples lacking explicit connections. Our results demonstrate that integrating ClinicalBERT node features and graph structure significantly improves classification performance, especially in fine-grained skill prediction. CFiCS achieves substantial gains in both micro and macro F1 scores across all tasks compared to baselines, including random forests, BERT-based multi-task models, and graph-based methods.
\end{abstract}

\section{Introduction}

Psychotherapy is a complex process that, despite diverse theories and techniques—from cognitive-behavioral and psychodynamic methods to humanistic approaches—shares common change principles that reinforce its effectiveness. These universal elements, known as common factors (CFs), include the therapeutic relationship, expectancy factors, corrective experiencing, insight, and self-efficacy \cite{bailey2023common} and account for around 30\% of therapy outcomes \cite{lambert-1992-psychotherapy}, with the therapeutic relationship being a particularly strong predictor of positive change \cite{nahum2019common}. Intertwined with these CFs are microcounseling skills—discrete, teachable behaviors introduced by \citet{ivey1968microcounseling}, such as reflective listening and the strategic use of open-ended questions. These skills enable therapists to evoke change principles in practice. For example, fostering the therapeutic bond (an element of the CF therapeutic relationship) by using reflective listening (a microcounseling skill) to convey an empathic and validating attitude (an intervention concept) will likely improve client involvement and therapeutic effectiveness.

Monitoring therapists' behaviors in relation to therapeutic processes can provide deeper insights into how these processes, in turn, contribute to improved treatment outcomes. Moreover, automating skill and CF identification in place of human-based coding or post-session questionnaires improves scalability, lowers costs, and enables the analysis of within-session micro-processes \cite{falkenstrom2017working}. This approach also offers targeted, session-by-session feedback, enabling clinicians to refine their techniques and adapt interventions to individual client needs. 

One effective way to structure change principles systematically is by modeling CFs and microcounseling skills as a graph-based taxonomy. In this taxonomy, CF elements serve as higher-level categories, while microcounseling skills act as specific subcategories or methods used to elicit these factors. The hierarchical relationships within the graph illustrate how particular skills are applied in the context of broader factors. Moreover, this graph can be further enriched by incorporating node attributes, such as detailed descriptions and contextual examples, that clarify how each skill functions within its corresponding CF.

Building on this structured representation, we can leverage graph machine learning (ML) models to classify text by encoding these relationships. These models can learn embeddings for each node, effectively capturing both the structural and feature-based information embedded in the taxonomy. For example, when analyzing a therapeutic interaction, the model can identify relevant microcounseling skills (like reflective listening or validation) and link them to higher-level CFs (such as the therapeutic bond).
Based on this framework, we propose a classification method 
\textit{CFiCS} \footnote{Code available on \href{https://github.com/smidtfab/CFiCS}{GitHub}} 
that employs graph ML to aggregate information from the interconnected network of \textbf{c}ommon \textbf{f}actors, \textbf{i}ntervention \textbf{c}oncepts, \textbf{s}kills, and examples. This approach allows us to inductively predict associations between previously unseen text and the corresponding CFs (e.g., therapeutic relationship), CF elements (e.g., therapeutic bond), intervention categories (e.g., collaboration and partnership), and skills (e.g., reflective listening), ultimately enhancing our ability to assess and improve therapeutic interactions. 
We demonstrate through experiments that integrating the graph outperforms baselines. The most accurate configuration combines ClinicalBERT embeddings with GraphSage. 

\section{Background and Related Work}

\subsection{Automatic Detection of Therapeutic Elements in Clinical Text}

The growing use of technology in psychotherapy has expanded the possibilities of automating text data collection, such as therapy transcripts. This has increased interest in using natural language processing and ML to automatically detect, classify, or score therapist behaviors and client responses. For instance, recent research has attempted to identify empathy-related cues in counseling dialogue \citep{tao24b_interspeech, tavabi2023therapist}. Other studies have focused on classifying types of reflections or questions posed by therapists \citep{can2016sounds,perez2017predicting}.
Despite this progress, several significant challenges remain. First, therapy transcripts inherently contain sensitive information, restricting the available data for model training. Second, counselor behaviors are highly context-dependent; the same microcounseling skill may evoke different CFs. For example, respect for the client's autonomy generally conveys an attitude of collaboration and partnership but is also an inherent part of goal alignment. Third, most existing studies focus on one specific behavioral construct (e.g., identifying therapist empathy alone) rather than a broad taxonomy encompassing multiple change principles and a wide range of microcounseling skills. 
Recent work explored fine-grained analysis of psychotherapy sessions. \citet{mayer-etal-2024-predicting} developed models that predict client emotions and therapist interventions at the utterance level. Similarly, \citet{gibson2019multi} introduced multi-label, multi-task deep learning methods that simultaneously predict multiple behavioral codes within therapy dialogues. Despite these advancements, most existing studies still emphasize isolated behavioral constructs (e.g., identifying therapist empathy alone) rather than addressing a broader taxonomy of multiple change principles and diverse microcounseling skills. These limitations motivate the need for more holistic, theory-driven computational approaches that can parse complex therapeutic interactions at multiple levels of granularity.

\subsection{Taxonomies and Graph-Based Modeling Approaches}

Researchers have explored structured representations like taxonomies or ontologies to capture the hierarchical and interconnected nature of therapeutic elements, such as broad CFs and more granular microcounseling skills, and have applied these frameworks to classify symptoms, diagnoses, and interventions in mental health research \citep{evans2021taxonomy}. However, few existing taxonomies systematically link higher-level CFs (e.g., the therapeutic relationship) to actionable skills (e.g., reflective listening, validating) that instantiate those factors in practice.
Graph-based ML offers a robust avenue for modeling these relationships. With methods such as Graph Convolutional Networks (GCNs) \citep{kipf2017semisupervised} or GraphSAGE \citep{hamilton2017inductive}, one can encode both the textual features of nodes (e.g., descriptions of skills) and the relational structure (e.g., which skills evoke which CFs) into a unified embedding space. Graph ML has shown promise in diverse classification tasks—among others, in detecting suicidality \cite{lee-etal-2022-detecting}—suggesting that a similar strategy could be applied to psychotherapy discourse.

\begin{figure*}[!t]
    \centering
    \includegraphics[width=1\linewidth]{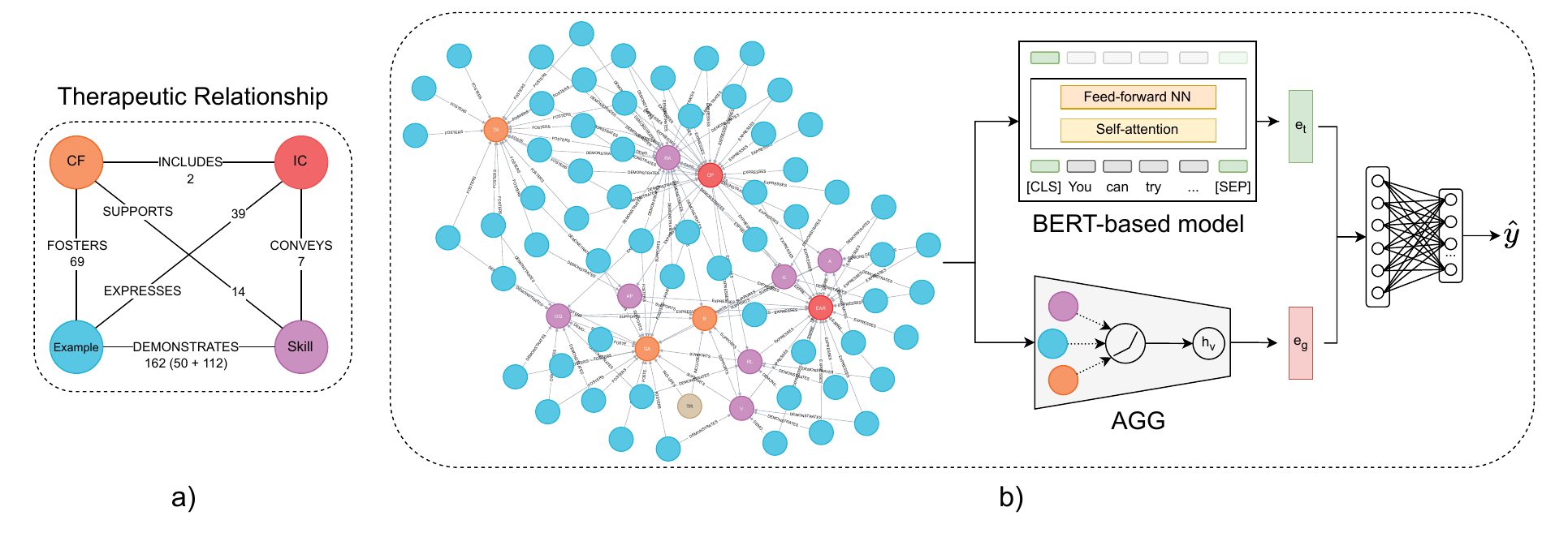}
    \caption{(a) Edge types and connection patterns between the node types for the common factor Therapeutic Relationship and (b)  Common factor and microcounseling skill prediction with our CFiCS classification model.}
    \label{fig:the_method}
\end{figure*}

\section{The CFiCS Graph}
We construct a structured knowledge graph of therapeutic practices centered around five main types of nodes: the root of our graph, the CF therapeutic alliance; the \textit{CF elements} (i.e., Bond, Goal Alignment, and Task Agreement), which define integral components of the CF therapeutic alliance; \textit{intervention concepts (ICs)} (i.e., Empathy, Acceptance, and Positive Regard; Collaboration and Partnership), which express specific approaches to fostering the CF elements; \textit{therapeutic skills} (i.e., Open-Ended Questions, Reflective Listening, Affirmation, Validation, Genuineness, Respect for Autonomy, Asking for Permission), which are practical techniques therapists use; and \textit{examples}, which are therapist statements that illustrate how specific skills and concepts are applied.
Figure~\ref{fig:the_method}a visualizes the connection patterns between the node types. The graph is heterogeneous, incorporating different node types and relationships, and sparse. Most examples link to one CF or one IC and one or two skills. Hence, the average path length is short due to the triadic pattern: example $\rightarrow$ skill $\rightarrow$ CF or IC. The structure is hierarchical, with CFs at the top, ICs as an intermediate layer, and therapeutic skills and examples forming the practical, grounded components. Clusters naturally form around specific CFs and ICs, creating thematic groupings.
The graph's relationships are multi-relational and include edges like \textit{fosters}, linking examples to the CFs they develop; \textit{expresses}, connecting skills to ICs or examples to concepts; and \textit{demonstrates}, linking examples to the therapeutic skills they showcase. This semantic structure provides a foundation for our classification approach.

\section{The CFiCS Classification Model}

We propose a model shown in Figure~\ref{fig:the_method}b to classify CFs, ICs, and \textit{associated} therapeutic skills for textual input. We leverage graph ML to exploit the topology of the nodes in the CFiCS graph and combine it with the textual embeddings produced by a pretrained language model. CFiCS enables inductive classification of previously unseen nodes, which do not have explicit edges but can still leverage the structural patterns learned from the graph during training in addition to the textual features.

\paragraph{Input graph structure}
Let $\mathcal{G} = (\mathcal{V}, \mathcal{E})$ be a heterogeneous graph where:
$\mathcal{V} = \mathcal{V}_r \cup \mathcal{V}_f \cup \mathcal{V}_c \cup \mathcal{V}_s \cup \mathcal{V}_e$ represents the set of all nodes, where $\mathcal{V}_r$ the root node, $\mathcal{V}_f$ the CF nodes, $\mathcal{V}_c$ the IC nodes, $\mathcal{V}_s$ the skill nodes, and $\mathcal{V}_e$ the example nodes.
$\mathcal{E} \subseteq {(v_i, v_j) | v_i, v_j \in \mathcal{V}}$ represents the bidirectional edges between nodes in the graph. There exist six distinct edge types, also visualized in Figure~\ref{fig:the_method}b:

\begin{enumerate}
    \item \textbf{Fosters relation}: \((v_{e}, v_{f})\) denotes a relationship between an example node and a CF node, indicating that the example fosters the development of a specific CF in therapy.

    \item \textbf{Expresses relation}: \((v_{e}, v_{c})\) connects an example node to an IC node, signifying that the example expresses the therapeutic intention of an IC (e.g., reflective listening expresses empathy, acceptance, and positive regard).

    \item \textbf{Demonstrates relation}: \((v_{e}, v_{s})\) links an example node to a skill node, showing that the example demonstrates the practical application of a specific therapeutic skill.

    \item \textbf{Includes relation}: \((v_{f}, v_{c})\) links a CF node to an IC node, indicating that the IC is a specific approach to operationalizing the broader CF.

    \item \textbf{Conveys relation}: \((v_{s}, v_{c})\) connects a skill node to an IC node, signifying that the skill conveys the therapeutic intention of an IC.

    \item \textbf{Supports relation}: \((v_{s}, v_{f})\) connects a skill node to a CF node, highlighting that a microcounseling skill supports a CF element.
    
\end{enumerate}

\paragraph{Node features}
For each node \( v \in \mathcal{V} \), let \( \tau_v \) denote the associated textual input. For nodes representing CFs, ICs, or skills, \( \tau_v \) consists of the node's name and detailed description. In contrast, for example nodes, \( \tau_v \) comprises solely the example text. A pretrained language model $\mathcal{M}$ (e.g., BERT \cite{devlin-etal-2019-bert} or ClinicalBERT \cite{alsentzer-etal-2019-publicly}) computes the feature vector for each node.
\begin{equation*}
\mathbf{x}_v = \mathcal{M}(\tau_v)
\end{equation*}
where \( \mathbf{x}_v \in \mathbb{R}^d \) and \( d \) is the embedding dimension of the chosen model (e.g., $d=768$ for BERT).



\paragraph{Learning task}
We seek to learn node embeddings via a \emph{message passing} framework inspired by GraphSAGE~\cite{hamilton2017inductive}. Given training node representations $\{\mathbf{x}_v \mid v \in \mathcal{V}\}$ and graph structure $\mathcal{E}$, we iteratively update each node $v$'s representation as
\begin{equation*}
    \mathbf{h}_v^{(l)} 
      = \gamma^{(l)}\!\Bigl(
          \mathbf{h}_v^{(l-1)},
          \mathrm{AGG}\!\bigl\{\mathbf{h}_u^{(l-1)} \mid u \in \mathcal{N}(v)\bigr\}
        \Bigr)
\end{equation*}
where $\mathbf{h}_v^{(l)}$ is $ v$'s embedding at layer $l$, $\gamma^{(l)}$ is a learnable transformation (often a nonlinear MLP), $\mathrm{AGG}$ is a neighborhood aggregation function (e.g., mean or pooling aggregator), and $\mathcal{N}(v)$ denotes $ v$'s neighbors. Although we present this in a GraphSAGE-oriented formulation, \emph{the same learning task is fully model-agnostic}: by substituting different forms of $\mathrm{AGG}$ (such as attention-based aggregation in GAT \cite{veličković2018graph} or weighted normalized sums in GCN \cite{kipf2017semisupervised}) and choosing a suitable $\gamma^{(l)}$, one can instantiate a variety of GNN variants without altering the underlying message passing framework.

\paragraph{Classification tasks}
Given an example node \(v \in \mathcal{V}_e\) with embedding \(\mathbf{h}_v\), we define a single linear classification layer that encompasses all labels (i.e., for CFs, ICs, and skills). Let \(\mathcal{T} = \{\text{CF}, \text{IC}, \text{Skill}\}\) denote the task types, and let \(\mathcal{V}_t\) be the set of labels for each task \(t \in \mathcal{T}\). We construct a single parameter matrix \(W\) and bias vector \(\mathbf{b}\) such that the row segments of \(W\) and the corresponding portions of \(\mathbf{b}\) map to the different tasks. Formally, the probability of assigning label \(y\) from \(\mathcal{V}_t\) to node \(v \in \mathcal{V}_e\) is computed by slicing the relevant portion of the linear output and applying a softmax.
\[
p_t(y \mid v) \;=\; \mathrm{softmax}\Bigl(\bigl(W_{\text{slice}(t)}\,\mathbf{h}_v + \mathbf{b}_{\text{slice}(t)}\bigr)\Bigr) 
\]
where $y \in \{1, \dots, |\mathcal{V}_t|\}$, \(\mathbf{h}_v\) is the shared node embedding (e.g., obtained from a graph neural network), \(W_{\text{slice}(t)}\) and \(\mathbf{b}_{\text{slice}(t)}\) refer to the rows in \(W\) and \(\mathbf{b}\) that correspond to the label set \(\mathcal{V}_t\), and \(\mathrm{softmax}(\cdot)\) is applied to the sliced logits to form a probability distribution specific to the task \(t\).

\paragraph{Optimization objective}
We treat each task $t \in \mathcal{T}$ as a separate multi-class classification problem and define a cross-entropy loss $\mathcal{L}_t$ on the predictions $p_t(y_v \mid v)$. Formally,
\[
    \mathcal{L}_t \;=\; - \sum_{v \in \mathcal{V}_e} \log\,p_t(y_v \mid v),
\]
where $y_v \in \mathcal{V}_t$ denotes the ground truth label for node $v$ in task $t$. Our overall multi-task objective is a linear combination of these losses
\[
    \mathcal{L}
    \;=\;
    \sum_{t \in \mathcal{T}} \lambda_t\,\mathcal{L}_t,
\]
with weights $\{\lambda_t\}$ controlling the relative importance of each task. Intuitively, each $\mathcal{L}_t$ measures how well the model performs on task $t$, and the hyperparameters $\lambda_t$ balance their contributions to the total loss.

\paragraph{Inference}

A new node is isolated during inference, meaning it has no edges and lacks direct neighbors in the graph. When the set of neighbors $\mathcal{N}(v)$ is empty, the aggregator defaults to relying solely on $h_v^{(l-1)}$. However, the model's learned weights still capture global patterns from the training graph. The aggregator, which processes node features, has learned the overall graph structure and label signals, allowing it to embed the new node in a graph-aware feature space. Even without access to neighbors, the aggregator's learned MLP transforms the new node's features with knowledge learned during training on graph edges.

\begin{table*}[ht]
\centering
\caption{Three-fold cross-validation micro and macro F1 scores.} 
\label{tab:f1_results}
\renewcommand{\arraystretch}{1.2} 
\resizebox{\textwidth}{!}{ 
\begin{tabular}{l|c@{\hskip 6pt}c@{\hskip 6pt}c|c@{\hskip 6pt}c@{\hskip 6pt}c}
\toprule
\multirow{2}{*}{\textbf{Model}} & \multicolumn{3}{c|}{\textbf{Micro F1}} & \multicolumn{3}{c}{\textbf{Macro F1}} \\
\cmidrule(lr){2-4}\cmidrule(lr){5-7}
 & CF & IC & Skill & CF & IC & Skill \\
\midrule
RF (TF-IDF multi-task)      
  & 52.50 $\pm$ 3.11   & 74.59 $\pm$ 1.48   & 53.02 $\pm$ 8.39  
  & 20.43 $\pm$ 3.82   & 38.04 $\pm$ 1.65   & 49.77 $\pm$ 4.24 \\
BERT (multi-task)           
  & 59.69 $\pm$ 4.95   & 79.02 $\pm$ 1.39   & 59.60 $\pm$ 8.97  
  & 34.34 $\pm$ 11.0  & 46.29 $\pm$ 6.84   & 59.23 $\pm$ 10.3 \\
\midrule
GCN without BERT            
  & 55.70 $\pm$ 5.77   & 70.06 $\pm$ 4.38   & 19.37 $\pm$ 4.63  
  & 17.86 $\pm$ 1.19   & 27.45 $\pm$ 1.00   & 4.04 $\pm$ 0.81   \\
GAT without BERT            
  & 56.89 $\pm$ 1.16   & 71.27 $\pm$ 2.42   & 23.24 $\pm$ 8.39  
  & 18.13 $\pm$ 0.24   & 27.74 $\pm$ 0.55   & 4.65 $\pm$ 1.34   \\
GraphSage without BERT      
  & 56.86 $\pm$ 2.08   & 70.04 $\pm$ 2.63   & 14.39 $\pm$ 1.27  
  & 20.19 $\pm$ 4.00   & 27.45 $\pm$ 0.60   & 3.14 $\pm$ 0.24   \\
\midrule
CFiCS GCN with ClinicalBERT 
  & 74.53 $\pm$ 16.62  & 86.30 $\pm$ 9.05   & 91.36 $\pm$ 14.97 
  & 66.06 $\pm$ 21.02  & 74.79 $\pm$ 17.69  & 88.63 $\pm$ 19.68 \\
CFiCS GAT with ClinicalBERT 
  & 91.98 $\pm$ 12.33  & 93.21 $\pm$ 10.2  & 93.83 $\pm$ 10.69 
  & 82.39 $\pm$ 27.45  & 89.35 $\pm$ 14.99  & 92.18 $\pm$ 13.55 \\
\midrule
CFiCS with ClinicalBERT (\textbf{ours})
  & \textbf{95.04} $\pm$ \textbf{7.00}   & \textbf{100.0} $\pm$ \textbf{0.00} & 96.30 $\pm$ 6.42  
  & \textbf{88.95} $\pm$ \textbf{16.14}  & \textbf{100.0} $\pm$ \textbf{0.00} & 96.09 $\pm$ 6.78  \\
CFiCS with BERT (\textbf{ours})           
  & \textbf{95.04} $\pm$ \textbf{5.33}   & 97.53 $\pm$ 4.28           & \textbf{96.89} $\pm$ \textbf{3.84}  
  & 84.24 $\pm$ 16.43           & 95.88 $\pm$ 7.13           & \textbf{97.14} $\pm$ \textbf{3.44}  \\
\bottomrule
\end{tabular}
}
\end{table*}

\section{Experiments}

\subsection{Implementation Details}
We use either \textit{bert-base-uncased} or \textit{ClinicalBERT} from Huggingface Transformers \cite{wolf-etal-2020-transformers} to encode the node name and description by average-pooling the last hidden state. 
We implement the model in Python using PyTorch Geometric. 
The model processes $2 \times 768$ input channels, 768 for the text embedding and 768 for the graph embedding, through hidden layers of 64 channels in a three-layer architecture with a 0.5 dropout rate. It handles three task components: three CFs, two ICs, seven skills, and a \textit{neutral} class for each task. During training, a forward pass computes logits for these tasks, followed by slicing the output into separate components and computing the task-specific losses weighted by predefined task weights (defaulting to uniform). The Adam optimizer \cite{kingma2017adam} updates the parameters, with training configured for up to 400 epochs, a learning rate of 1e-3, and a weight decay of 1e-4. The model performs validation at each epoch, tracks the lowest loss, and stops early if it detects no improvement for 50 epochs.

\subsection{Dataset}
Our dataset consists solely of manually created and curated examples drawn from established psychotherapy literature, rather than real patient conversations.
The dataset is structured as a heterogeneous, undirected, and multi-relational CFiCS graph modeling therapeutic practices. It includes three CF elements (e.g., Bond), two ICs (e.g., Collaboration and Partnership), and seven therapeutic skills (e.g., Reflective Listening). The dataset contains 69 fully annotated examples, including CF, IC, and skill annotations, and an additional 112 examples annotated only for therapeutic skills, illustrating their application contextually. An expert selected and curated these examples from respected psychotherapy literature on therapeutic alliance. Specifically, we identified reference samples directly from \citet{fuertes2019working, miller2021effective, bailey2023common} as representative instances of therapeutic interaction for each class. These original excerpts served as a reference for generating new synthetic samples using ChatGPT, ensuring alignment with the themes, styles, and therapeutic concepts illustrated in the literature.
Examples not assigned to any class are designated as \textit{neutral}. The dataset is split into training and testing subsets (80/20), with k-fold cross-validation applied to the training data for model evaluation.

\subsection{Baselines}
We compare our approach against two baselines: a Random Forest (RF) \cite{breiman2001random} and a BERT-based architecture. Additionally, we evaluate and compare several graph ML methods, including GAT, GCN, and GraphSage, to assess their effectiveness in modeling the relationships within the CFiCS graph. For the RF baseline, we convert each utterance into TF-IDF features and create a multi-output target vector where CFs, ICs, and skills are multi-class tasks. We then train a \texttt{MultiOutputClassifier}, effectively training one RF per output dimension. For the BERT-based model, we finetune a pretrained encoder that extracts a pooled \texttt{[CLS]} representation and optimizes three classification heads using cross-entropy loss for the CF, IC, and skill prediction.


\subsection{Metrics}
We report macro- and micro-averaged F1 scores for the multi-class tasks.
The macro F1 treats each class equally, computing the mean F1 over classes, whereas the micro F1 aggregates contributions from all classes to compute precision and recall overall. The micro F1 tallies the total number of correctly predicted skill labels versus all predictions, while the macro F1 averages the F1 values per skill category.
In addition, we use Precision@k and Recall@k as indicators of cluster quality to evaluate how well the model ranks and groups related concepts. Precision@k measures the proportion of relevant samples among the top-k similar nodes, reflecting how accurately the model embedds input features. Recall@k quantifies the proportion of relevant samples retrieved within the top-k similar samples, indicating how well the model captures relevant clusters. Higher Precision@k and Recall@k scores suggest that similar concepts are embedded closely together, providing additional validation of the quality of learned representations.

\section{Results}

\subsection{Quantitative Results}

\begin{table*}[ht]
    \centering
    \caption{Precision@\,$k$ and Recall@\,$k$ for GAT, GCN, and GraphSage with ClinicalBERT ($k \in \{1,5,10\}$).} 
    \label{tab:simplified_k1_k5_k10}
    \begin{tabular}{llcccccc}
        \toprule
        \textbf{Class} & \textbf{Model} & \textbf{P@1} & \textbf{R@1} & \textbf{P@5} & \textbf{R@5} & \textbf{P@10} & \textbf{R@10} \\
        \midrule
        \multirow{3}{*}{Common Factors}
        & GAT        
            & \textbf{10.81} & \textbf{1.03} 
            & 57.84 & 26.46
            & \textbf{63.78} & 50.46 \\
        & GCN        
            & 0.00 & 0.00
            & 45.41 & 24.04
            & 46.22 & 41.80 \\
        & GraphSage  
            & 0.00 & 0.00
            & \textbf{58.92} & \textbf{33.47}
            & 59.73 & \textbf{57.96} \\
        \midrule
        \multirow{3}{*}{Intervention Concepts}
        & GAT        
            & \textbf{13.51} & \textbf{1.29}
            & 71.35 & 28.83
            & 74.32 & 45.85 \\
        & GCN        
            & 0.00 & 0.00
            & 62.16 & 20.47
            & 64.05 & 35.44 \\
        & GraphSage  
            & 0.00 & 0.00
            & \textbf{76.76} & \textbf{32.82}
            & \textbf{77.03} & \textbf{46.62} \\
        \midrule
        \multirow{3}{*}{Skills}
        & GAT        
            & \textbf{13.51} & \textbf{3.60}
            & 55.14 & \textbf{71.17}
            & 29.19 & 76.13 \\
        & GCN        
            & 0.00 & 0.00
            & 48.65 & 64.11
            & 32.97 & \textbf{77.18} \\
        & GraphSage  
            & 0.00 & 0.00
            & \textbf{57.30} & 55.08
            & \textbf{37.84} & 71.98 \\
        \bottomrule
    \end{tabular}
\end{table*}

\begin{figure}
    \centering
    \includegraphics[width=0.97\linewidth]{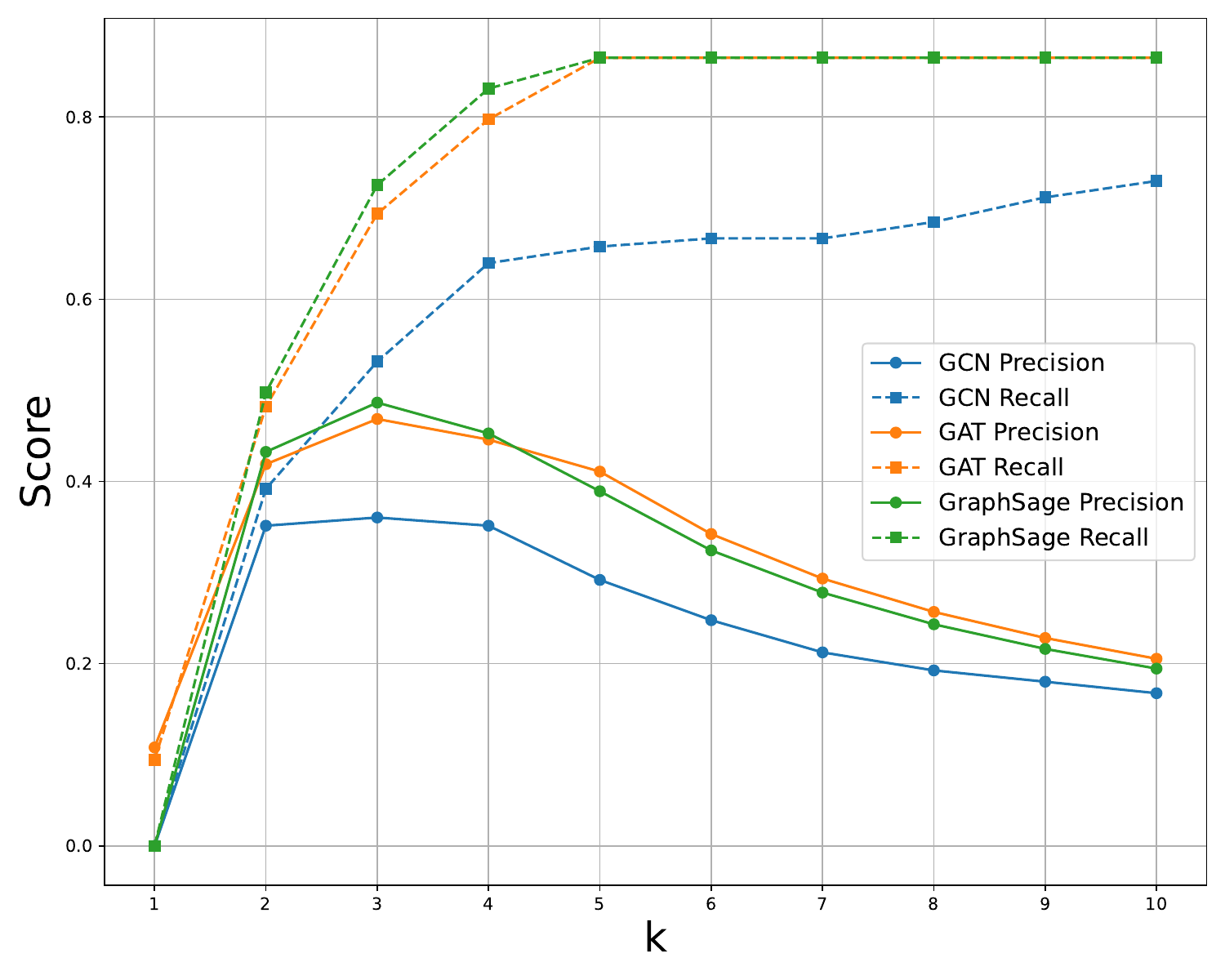}
    \caption{Precision and recall at different values of  $k$  (ranging from 1 to 10) for graph-based models, including GCN, GAT, and GraphSage.}
    \label{fig:prec_rec@k}
\end{figure}

\paragraph{Performance of non-graph baselines}

Table~\ref{tab:f1_results} reports the micro and macro F1 scores for various models on the CF, IC, and skill classification. We compare several baselines, including traditional TF-IDF and BERT multi-task classifiers, graph-based models without BERT, and our proposed CFiCS variants that integrate clinical BERT features.
The TF-IDF multi-task Random Forest baseline achieves relatively modest performance, with micro F1 scores of 52.50, 74.59, and 53.02 for CF, IC, and Skill, respectively, and corresponding macro F1 scores of 20.43, 38.04, and 49.77. The BERT multi-task classifier improves these numbers considerably (e.g., obtaining 59.69 and 59.60 micro F1 for CF and Skill, respectively), indicating the benefit of richer contextual representations.

\paragraph{Effect of graph structure on performance}

Graph-based models without BERT features exhibit mixed results. For example, the GCN and GAT models without BERT yield micro F1 scores in the range of 55.70--56.89 for CF and around 70 for IC. However, their performance on the skill task is markedly lower (with micro F1 scores of 19.37, 23.24, and 14.39 for GCN, GAT, and GraphSage, respectively). This suggests that relying solely on graph structure without contextual text representations can be detrimental, particularly for the more nuanced skill classification.
In contrast, our proposed methods that incorporate clinical BERT features within the CFiCS framework demonstrate substantial improvements. Both the CFiCS GCN and CFiCS GAT variants with ClinicalBERT improve performance across all tasks. In particular, the CFiCS GAT with ClinicalBERT variant achieves micro F1 scores of 91.98, 93.21, and 93.83 for CF, IC, and Skill, respectively, with corresponding macro F1 scores of 82.39, 89.35, and 92.18.
The CFiCS models that integrate BERT features achieve the best results. The model labeled as \emph{CFiCS with ClinicalBERT (ours)} achieves a micro F1 of 95.04 on CF and 96.30 on Skill, with perfect performance on the IC task (100.00 in both micro and macro F1). Similarly, \emph{CFiCS with BERT (ours)} shows competitive performance with micro F1 scores exceeding 95\% for CF, IC, and Skill, and macro F1 scores that are consistently high.

\begin{figure*}
    \centering
    \includegraphics[width=1\textwidth]{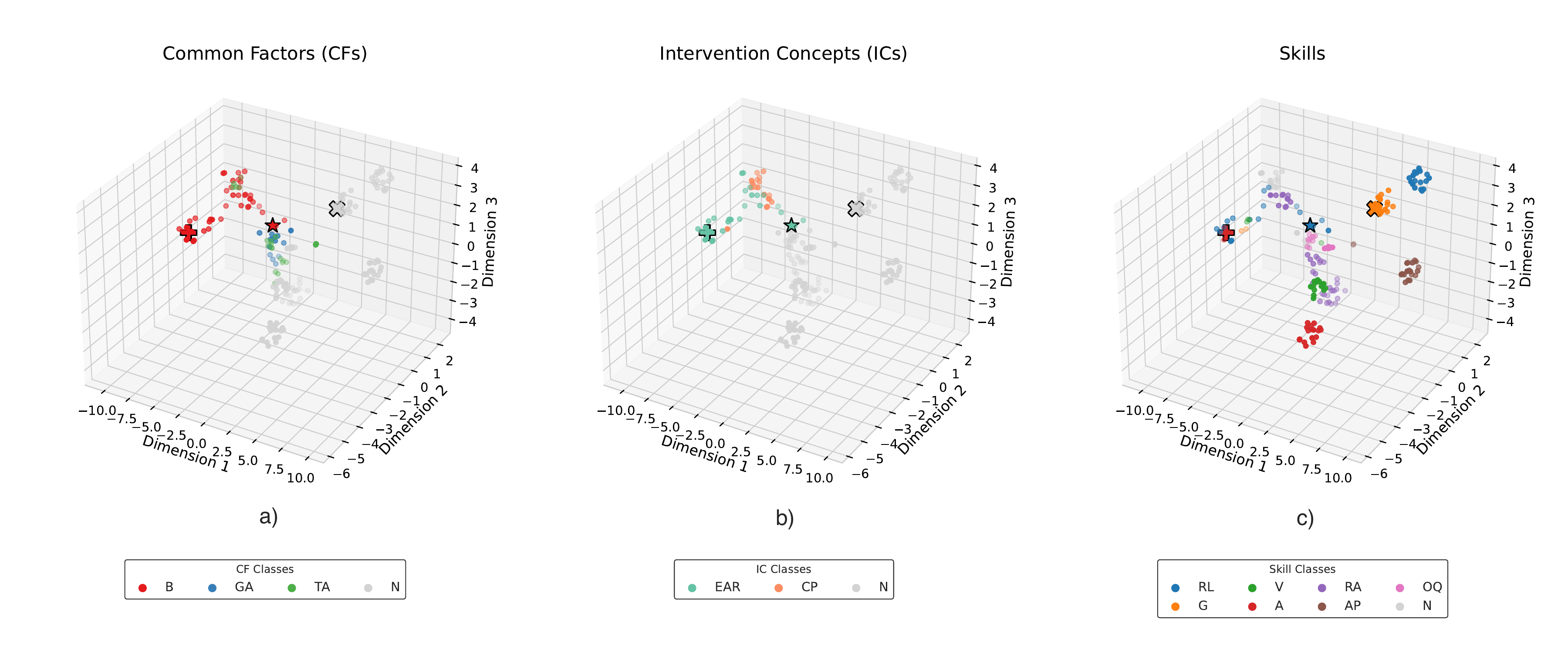}
    \caption{Visualization of the 197 learned embeddings after aggregating contextual information from the graph. TSNE reduces the 832-dimensional vectors (768 text features + 64 graph features) to two dimensions. Each point represents a node embedding, with colors indicating the three different class types. The embeddings are extracted from the final hidden layer during the forward pass.}
    \label{fig:embeddings_visualized}
\end{figure*}

\paragraph{Precision and recall at different k}
We evaluate the embedding quality by comparing the precision and recall at varying $k$ ranging from one to ten. 
Figure~\ref{fig:prec_rec@k} presents the aggregated precision and recall for values of k  ranging from 1 to 10. As expected, recall increases as k grows since more relevant items are retrieved, while precision declines due to the broader set of top-k retrievals. Table~\ref{tab:simplified_k1_k5_k10} provides a more detailed breakdown of precision and recall across different class types.
GraphSage demonstrates superior recall and precision at higher k, suggesting that its embeddings create more cohesive clusters of relevant nodes, making it more effective for retrieving multiple correct labels. GCN lags behind both models in rank-1 precision but improves recall at k=10. While its embeddings do not strongly differentiate the best match, they still capture helpful information for broader retrieval. Overall, GAT is best for fine-grained differentiation, GraphSage generates well-structured clusters that enhance overall representation quality, and GCN provides moderate performance with embeddings that favor broader contextual generalization.

\subsection{Qualitative Results}

In Figure~\ref{fig:embeddings_visualized}, we visualize the learned embeddings in 3-dimensional space using the dimension reduction method t-SNE \cite{vandermaaten2008tsne}. We color the samples based on their respective class for all three class types (CF, IC, and skill). 
We observe that skills form well-separated clusters, indicating that the model effectively distinguishes between different microcounseling techniques. 
Additionally, higher-level CF influence representation structure, as utterances containing a skill alone (e.g., Genuineness, \ding{54} in Figure~\ref{fig:embeddings_visualized}c) are embedded separately from those where the skill co-occurs with broader therapeutic elements (e.g., Bond (B) or Empathy, Acceptance, and Positive Regard (EAR) (see the \ding{73} and \ding{58} sign in Figure~\ref{fig:embeddings_visualized}a and b). This suggests that the model captures fine-grained skill differentiation and hierarchical relationships between skills and ICs. Moreover, higher-level features such as the CF and IC cluster more closely together, showing less separation. The context of the skills separates them at the lower, more fine-grained skill level. This finding indicates that while the model captures distinct skill representations, the broader context in which practitioners apply these skills is crucial for differentiation.


\section{Ethical and Impact Considerations}
Psychotherapy transcripts contain highly sensitive and personal information, and patients are particularly vulnerable data subjects. However, our dataset consists solely of manually created and curated examples from the literature, not real patient conversations, ensuring no private or identifiable data is used. Using this method with real-world therapy data would require strict attention to privacy and confidentiality, ensuring compliance with data protection regulations.
ML models trained on limited or biased datasets can inadvertently learn and propagate biases in the data. Since our dataset is relatively small and manually curated, there is a risk that certain features are over- or underrepresented, potentially impacting the generalizability of our results.
Furthermore, automated classification of psychotherapy content could be misused if applied without proper oversight. For instance, misclassification of therapeutic interactions could lead to inaccurate feedback for therapists, and reliance on imperfect AI-driven assessments might undermine professional judgment. Therefore, the model should be deployed as an assistive tool rather than replacing human expertise.

Our work has implications for both psychotherapy research and practice. 
Firstly, traditional research methods often fail to capture the complexity of the patient-therapist interaction process \citep{lundh2019towards}. For example, studies on therapeutic alliance patterns typically rely on post-session evaluations, which may oversimplify evolving patient-therapist dynamics \citep{falkenstrom2017working}. Automatic assessment of in-session "microprocesses" \citep{lundh2019towards} could offer a more precise understanding of common factors development, identifying key therapist skills linked to treatment success across modalities and client profiles.

Secondly, psychotherapy quality in practice depends on research-driven training and performance-based feedback \citep{baldwin2013therapist}. Yet, many clinicians receive little feedback after initial training \citep{moyers2005assessing}. A system providing session-by-session feedback on common factor usage on various levels of granularity could help therapists set incremental improvement goals and track progress in real time \citep{rousmaniere2016deliberate}. 
Thirdly, automating common factor feedback would enable integration with digital health tools, linking therapist skill use to broader treatment data, including symptom levels and session attendance.

\section{Conclusion}

We presented a graph ML classification method \textit{CFiCS} to classify common factors, intervention concepts, and associated skill usage.
Overall, the results demonstrate that combining textual features from ClinicalBERT with graph-based ML in the CFiCS framework significantly enhances classification performance, particularly for the challenging skill prediction task, and outperforms conventional TF-IDF, BERT, and pure graph-based baselines.



\section{Limitations}
One primary limitation is the dataset size. We evaluate our method using a manually curated dataset alongside examples from the literature. While the model demonstrates promising performance, the relatively small sample size may limit generalizability and increase the risk of overfitting to specific linguistic patterns or annotation biases.
A second limitation is language dependence. Our study focuses exclusively on English-language data, and we do not assess whether the method generalizes to other languages or multilingual settings. Given that therapeutic discourse varies linguistically and culturally, future work should explore cross-lingual adaptations and assess whether pretrained multilingual models (e.g., XLM-R, mBERT) can extend classification performance to other languages. An additional challenge is linguistic ambiguity per se. Identical statements can have different meanings depending on the context. Prosodic features play a key role in language comprehension \cite{dahan2015prosody}, and models trained on spoken language outperform text-based approaches \citep{singla2020towards}. Thus, CFiCS classification could benefit from incorporating auditory and visual cues.
Another limitation is the lack of external validation on out-of-distribution datasets. Our dataset consists solely of manually curated literature examples and synthetically generated examples, rather than real therapist-patient interactions. While this approach has ethical advantages by avoiding privacy concerns, it limits the clinical relevance of the dataset. 
Additionally, the effectiveness of common factor usage depends on their thoughtful application rather than mere frequency. Therapist responsiveness, seen as a “metacompetency” integrating skills like executive functioning and reflection \citep{hatcher2015interpersonal}, is more valuable than rigid technique use \citep{stiles2017appropriate}.

\section{Future Work}

Future work should expand the dataset and use real therapy interactions in different settings and with therapists using different approaches. Additionally, it may be beneficial to explore multilingual extensions, expand the CFiCS graph structure, validate on external corpora, and consider integrating prosodic features.

\section*{Acknowledgments}
This work is funded by the Research Council of Norway (grant 321561) through the Adaptive Level of Effective and Continuous Care (ALEC2) project and was partially supported by the Wallenberg AI, Autonomous Systems and Software Program (WASP), funded by the Knut and Alice Wallenberg Foundation.
The model training was enabled by the Berzelius resource provided by the Knut and Alice Wallenberg Foundation at the National Supercomputer Centre.

\bibliography{anthology,custom}




\end{document}